\documentclass[pdflatex,sn-mathphys-num]{sn-jnl}

\usepackage{graphicx}%
\usepackage{multirow}%
\usepackage{amsmath,amssymb,amsfonts}%
\usepackage{amsthm}%
\usepackage{mathrsfs}%
\usepackage[title]{appendix}%
\usepackage{xcolor}%
\usepackage{textcomp}%
\usepackage{manyfoot}%
\usepackage{booktabs}%
\usepackage{algorithm}%
\usepackage{algorithmicx}%
\usepackage{algpseudocode}%
\usepackage{listings}%
\usepackage{CJKutf8}%
\usepackage[numbers,sort&compress]{natbib}
\setcitestyle{super} 

\theoremstyle{thmstyleone}%
\theoremstyle{thmstyletwo}%

\theoremstyle{thmstylethree}%

\begin{document}

\title{Advancing multi-site emission control: A physics-informed transfer learning framework with mixture of experts for carbon-pollutant synergy}


\author[1]{\fnm{Yuxuan} \sur{Ying}}\email{yuxuanying@zjut.edu.cn}
\equalcont{These authors contributed equally to this work.}

\author[2]{\fnm{Hanqing} \sur{Yang}}\email{yhq1778@gmail.com}
\equalcont{These authors contributed equally to this work.} 

\author[3]{\fnm{Kaige} \sur{Wang}}\email{17596139731@163.com}

\author[4]{\fnm{Yu} \sur{Hu}}\email{yuhu2000@zjut.edu.cn}

\author[3]{\fnm{Zhiming} \sur{Zheng}}\email{zzm120229@163.com}

\author[3]{\fnm{Yunliang} \sur{Jiang}}\email{jyl2022@zjnu.cn}

\author[5]{\fnm{Xiaoqing} \sur{Lin}}\email{linxiaoqing@zju.edu.cn}

\author[5]{\fnm{Xiaodong} \sur{Li}}\email{lixd@zju.edu.cn}

\author*[3]{\fnm{Jun} \sur{Chen}}\email{junc.change@zjnu.edu.cn}

\affil[1]{\orgdiv{College of Mechanical Engineering}, \orgname{Zhejiang University of Technology}, \orgaddress{\city{Hangzhou}, \postcode{310023}, \country{China}}}

\affil[2]{\orgname{Alibaba Group}, \orgaddress{\city{Hangzhou}, \postcode{311100}, \country{China}}}

\affil*[3]{\orgdiv{School of Computer Science and Technology}, \orgname{Zhejiang Normal University}, \orgaddress{\city{Hangzhou}, \postcode{311231}, \country{China}}}

\affil[4]{\orgdiv{Science and Education Integration College of Energy and Carbon Neutralization}, \orgname{Zhejiang University of Technology}, \orgaddress{\city{Hangzhou}, \postcode{310023}, \country{China}}}

\affil[5]{\orgdiv{State Key Laboratory of Clean Energy Utilization}, \orgname{Zhejiang University}, \orgaddress{\city{Hangzhou}, \postcode{310027}, \country{China}}}


\abstract{Municipal solid waste incineration (MSWI) converts urban waste to energy but simultaneously emits carbon dioxide, carbon monoxide and multiple regulated air pollutants whose formation is tightly coupled within a single combustion system. Controlling these emissions across a network of diverse facilities poses a fundamentally different challenge from optimising a single plant: data-driven models trained at one site capture local statistical patterns that rarely survive transfer to another, because they lack the physical constraints and regime-level structure needed to generalise. Here we show that shared emission--control relationships can be identified across heterogeneous MSWI plants when physical conservation laws, operating-regime heterogeneity and carbon--pollutant coupling are treated jointly. We develop a carbon--pollutant mixture-of-experts (CPMoE) model that routes process observations through regime-specific expert networks under conservation-based regularisation, and combine it with physics-informed transfer learning to adapt a reference model to new facilities. Across 13 plants, CPMoE predicts six major pollutants and a composite system-level risk index with source-domain $R^2$ of 0.668--0.904 and 0.666--0.970, respectively; after transfer to 12 target plants these values remain 0.661--0.842 and 0.610--0.841. Expert-utilisation patterns show that adaptation proceeds through structured regime re-weighting rather than re-learning from scratch. Embedding the transferred model in an offline digital twin and screening candidate operating adjustments against historical process records yields consistent risk-index reductions of 3.6--6.3\% with simultaneous pollutant co-reductions in 94--100\% of evaluated samples. These findings suggest a practical route toward transferable, system-level decision support for carbon--pollutant co-control in heterogeneous waste-to-energy networks.}

\keywords{Waste-to-energy systems, carbon--pollutant synergy, physics-informed machine learning, transfer learning, mixture of experts, digital twin}



\maketitle

\section{Introduction}\label{sec1}

Municipal solid waste incineration (MSWI) has become an important component of modern waste management systems, particularly in densely populated regions where landfill capacity is limited \cite{he2022global,zhang2022carbon}. Beyond waste reduction, MSWI operates as a waste-to-energy process in which carbon conversion, pollutant formation and energy recovery are closely coupled \cite{liu2025refocusing,tang2023air}. Incineration contributes to energy supply and landfill diversion, but it also remains a persistent source of atmospheric emissions---carbon dioxide ($\mathrm{CO}_2$), carbon monoxide ($\mathrm{CO}$), nitrogen oxides ($\mathrm{NO}_{\mathrm{x}}$), sulfur dioxide ($\mathrm{SO}_2$), hydrogen chloride ($\mathrm{HCl}$) and particulate matter ($\mathrm{PM}$)---that differ widely in formation pathways and environmental consequences \cite{tang2023air,ma2024decreasing,fei2025new}. Because all these pollutants are produced and controlled within the same combustion system, emission management is inherently a coupled operational problem rather than a collection of independent compliance tasks.

A major difficulty is that MSWI plants are highly heterogeneous \cite{ding2023dynamic,cui2024multi}. Differences in waste composition, furnace configuration, air supply strategy, operating conditions and pollution-control technology create substantial variability in emission behaviour---between plants, and even between furnaces within the same facility \cite{yan2024knowledge,qi2025novel}. An operating adjustment that reduces emissions in one incinerator may be ineffective or counterproductive in another \cite{xu2026universal}. Operators therefore tend to rely on site-specific empirical tuning, and effective emission-control practices rarely transfer across the expanding waste-to-energy network.

The dominant modelling approach has been to treat each pollutant as an independent prediction target \cite{wang2024modular,zhang2024heterogeneous,ma2024machine,wang2024c02}. Statistical, machine learning and deep learning models trained on single-plant operational data have improved within-plant prediction accuracy in many cases \cite{ding2023gradient,ma2024numerical,yang2023predicting}, but their generalisation to new sites is typically poor \cite{tian2022combustion,lee2024artificial,kabugo2020industry}. More fundamentally, single-pollutant modelling severs the mechanistic links that connect emissions: combustion efficiency, excess air, temperature and fuel characteristics jointly shape the full pollutant spectrum \cite{lin2025collaborative,feng2026unveiling}, and optimising one species in isolation can adversely affect others. This makes single-pollutant models ill-suited for reasoning about carbon--pollutant trade-offs or for supporting system-level decision-making \cite{sun2023data,huang2025cooperative}.

Increasing model complexity or expanding training datasets is unlikely to resolve this fundamental tension. Purely data-driven models learn site-specific patterns that support local interpolation but are fragile under domain shift \cite{zhuang2020comprehensive,azari2023systematic}. The deeper question is whether heterogeneous MSWI plants share transferable emission--control structures---relationships among combustion state, multi-pollutant formation and control response that remain meaningful across differences in plant configuration, operating regime and measurement baseline. Without such shared structure, a model trained at one facility has no principled basis for transfer to another, and the broader goal of network-level, carbon--pollutant co-control remains out of reach.

Physics-informed machine learning provides a principled route to the kind of structural constraints that purely data-driven approaches lack, by embedding conservation laws and process knowledge directly into the learning objective \cite{zhu2024physics}. Physics-informed neural networks have demonstrated improved stability and generalisation in several dynamical systems by anchoring learned representations to physically admissible solutions. However, most existing applications target single-output prediction tasks or relatively well-defined physical systems; their applicability to complex heterogeneous industrial processes remains limited \cite{faiz2025optimizing,adi2026municipal}. MSWI adds a further complication: distinct combustion modes and control strategies operate under different conditions, so a single global emission--control mapping cannot adequately represent the full process space \cite{taassob2024pinn,zhang2024crk}.

Recent advances in deep transfer learning have begun to address cross-domain modelling in industrial settings, but important gaps remain \cite{maschler2021deep,yadav2024investigation}. Models can still latch onto site-specific correlations that break down when conditions shift or when a model is applied to a new facility. The key limitation is not only a lack of data, but also the absence of structural constraints that keep multi-pollutant predictions physically coherent across sites \cite{qiao2024dynamic,chen2024co}. A practical framework for multi-site MSWI modelling therefore needs to represent multiple operating regimes explicitly, enforce physically meaningful relationships during cross-site adaptation, and evaluate carbon and pollutant outcomes through a common system-level lens \cite{koldasbayeva2024challenges}.

\begin{figure*}[t]
\centering
\includegraphics[width=\textwidth]{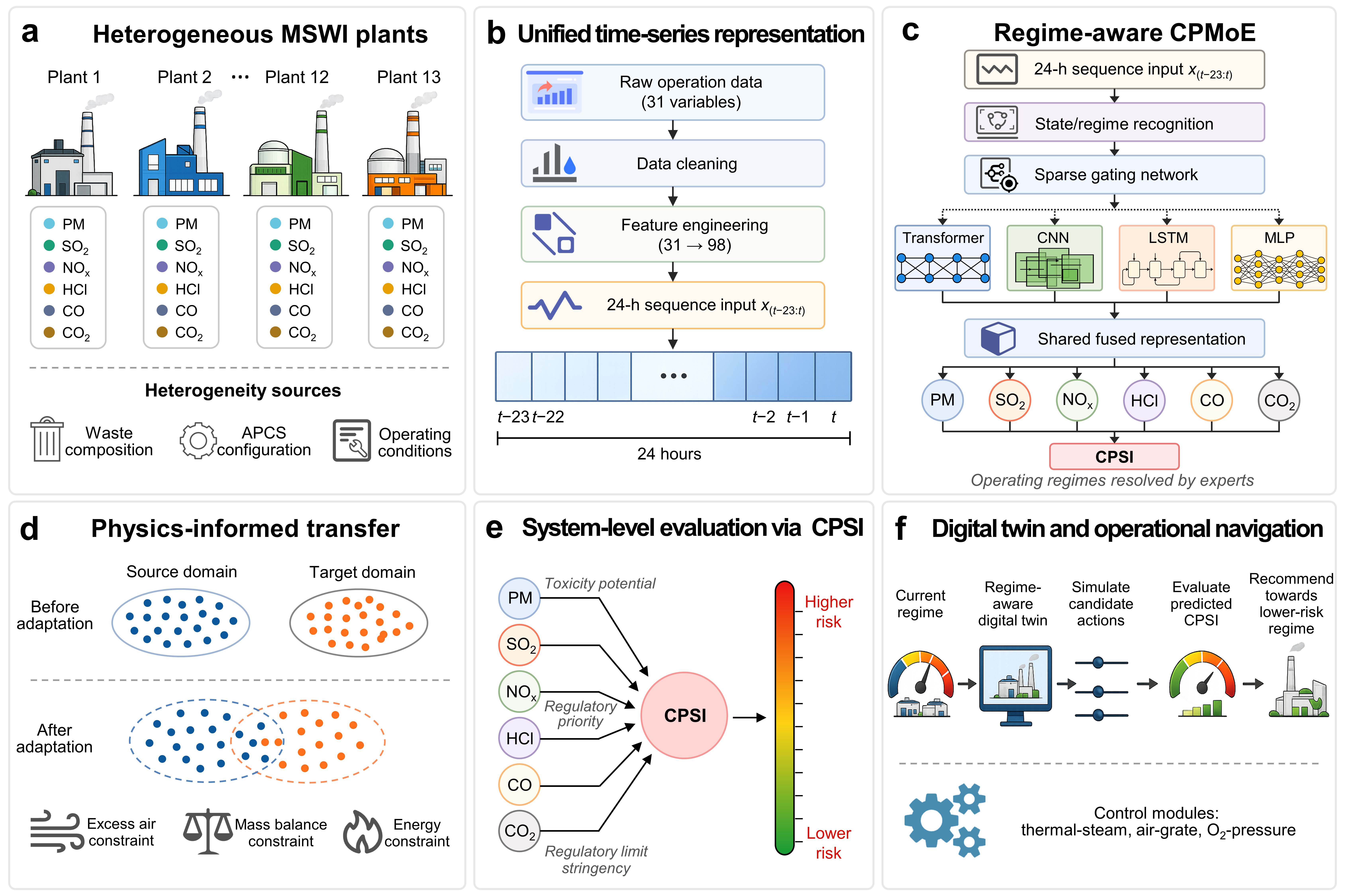}
\caption{
\textbf{Framework for multi-site MSWI emission modelling and control-oriented navigation.}
(a) Heterogeneous MSWI plants produce coupled carbon--pollutant emissions under differences in waste composition, APCS configuration and operating conditions.
(b) Raw operational variables are converted into a unified time-series representation through data cleaning, feature engineering and 24-hour sequence construction.
(c) The regime-aware CPMoE model represents multiple operating regimes using state recognition, sparse gating and heterogeneous experts (Transformer, CNN, LSTM and MLP), and produces multi-pollutant predictions that are aggregated into CPSI.
(d) Physics-informed transfer learning aligns source and target plants in a mechanism-consistent latent space using excess-air, pollutant-conservation and energy-balance constraints.
(e) Predicted pollutant concentrations are integrated into the carbon--pollutant synergistic index (CPSI) for system-level comparison of operating states.
(f) The learned representation is used in an offline digital twin, where candidate control actions are evaluated to identify operating directions associated with lower CPSI.
}
\label{fig:0}
\end{figure*}

In this study, we reframe MSWI emission modelling as the identification of transferable emission--control structures in heterogeneous, physics-constrained environments (\textbf{Fig.}~\ref{fig:0}). Rather than building plant-specific predictors, we ask whether shared relationships among combustion state, multi-pollutant formation and integrated emission risk can be learned from data across facilities. We develop a physics-informed mixture-of-experts model---the carbon--pollutant mixture-of-experts (CPMoE)---that combines conservation-based regularisation with data-driven representation learning, allowing distinct operating regimes to be identified within and across plants. A carbon--pollutant synergistic index (CPSI) is introduced to compare operating states in terms of integrated carbon--pollutant risk and to track whether system-level emission structure is preserved through transfer and operational navigation.

\section{Methods}\label{sec5}

\subsection{Problem Definition and Framework}

We formulate MSWI emission modelling as a system identification problem rather than a standalone prediction task. The aim is to learn emission--control relationships that remain meaningful under heterogeneous operating conditions and can therefore support cross-site transfer. For each sample, the model takes a sequence of process measurements over the previous $T$ time steps. These measurements include 31 original process variables, such as temperature, pressure, flow rate and oxygen concentration. The prediction targets are the concentrations of six major emissions: $\mathrm{PM}$, $\mathrm{SO}_2$, $\mathrm{NO}_{\mathrm{x}}$, $\mathrm{HCl}$, $\mathrm{CO}$ and $\mathrm{CO}_2$. The framework maps raw process histories to pollutant concentrations and then to system-level carbon--pollutant evaluation through CPSI. The overall architecture is shown in \textbf{Fig.} \ref{fig:0}b--d.

\subsection{Data Preprocessing}

Data preprocessing and feature engineering are designed to preserve physically meaningful temporal patterns in MSWI operation, including medium-term thermal inertia, load fluctuations, and diurnal operational cycles, rather than solely optimizing statistical properties. We adopt a triple imputation strategy to handle missing values. First, forward filling is applied to replace missing values with the previous valid observation. Missing values at the beginning of the sequence are then handled by backward filling. Finally, any remaining missing values are filled with zeros. For outlier treatment, considering the significant differences in data distribution characteristics among different pollutants, a per‑indicator adaptive percentile cleaning method is employed. $\mathrm{PM}$ and $\mathrm{HCl}$ data are thresholded using the 5th–95th percentiles, while $\mathrm{NO}_{\mathrm{x}}$ data are thresholded using the 2nd–98th percentiles. CO data exhibit a highly right‑skewed distribution with a large number of zero values, requiring special treatment. We also design an intelligent zero‑value processing strategy to distinguish between sensor‑fault zeros and true low‑value zeros. For fault zeros that appear consecutively three or more times, linear interpolation is applied for repair. Isolated zeros are replaced by small random numbers to avoid disrupting the natural distribution characteristics of the data.

To enhance the model’s ability to capture temporal patterns, systematic feature expansion is performed. Based on the original 31‑dimensional features, a 6‑hour sliding‑window moving average feature is added to capture medium‑term trends. First‑order difference features are introduced to characterize rate of change and fluctuation information. Hour‑of‑day and day‑of‑week cyclical encoding features are incorporated to capture periodic patterns. In addition, interaction features of key process parameters are included. After feature engineering, the input feature dimension is expanded from 31 to 98. Given the distribution differences among features, we propose an adaptive data transformation strategy. According to the skewness of each feature, an appropriate transformation method is automatically selected. For features with skewness greater than 1 (highly skewed), logarithmic transformation is applied. For features with skewness between 0.5 and 1 (moderately skewed), square‑root transformation is used. Features with skewness less than 0.5 (low skewness) remain unchanged. After transformation, Z‑score standardization is performed to bring all features to a common scale.

\subsection{Carbon-Pollutant Mixture-of-Experts (CPMoE)}

MSWI systems operate under multiple combustion and control regimes, each characterized by distinct emission--control relationships. A single global mapping is therefore insufficient to represent the system behaviour across the full operating space. To explicitly represent this regime heterogeneity, we adopted a Carbon--Pollutant Mixture-of-Experts (CPMoE) architecture \cite{rajbhandari2022deepspeed}, in which each expert captures a regime-consistent emission-control mechanism. In this model, different experts capture complementary temporal or operational patterns, while a gating network determines their sample-specific contributions. The overall architecture of CPMoE comprises four core components: a process state recognition module, a gating network, a heterogeneous expert system, and a multi-task output layer.

The process state recognition module initially processes the input time-series data $X$. This module extracts local temporal features through a one-dimensional convolutional layer, followed by layer normalization:
\begin{equation}
    H_{state}=\operatorname{LayerNorm}(\operatorname{Conv1D}(X)) .
\end{equation}
The normalized features are then fed into a bidirectional gated recurrent unit (BiGRU) to capture forward and backward temporal dependencies:
\begin{equation}
    \overrightarrow{h_t}, \overleftarrow{h_t} =\operatorname{BiGRU}(H_{state}),
\end{equation}
where $\overrightarrow{h_t}$ and $\overleftarrow{h_t}$ denote the hidden states of the forward and backward GRUs at time step $t$, respectively. To aggregate information across the entire sequence, an attention mechanism is introduced to compute the importance weights of each time step: 
\begin{equation}
    \alpha_t=\operatorname{softmax}(W^T_a \operatorname{tanh}(W_q[\overrightarrow{h_t}|| \overleftarrow{h_t}])),
\end{equation}
where $W_a$ and $W_q$ are learnable parameters, and the symbol $||$ represents the vector concatenation operation. The weighted sum based on attention weights yields the overall representation of the sequence: 
\begin{equation}
    z=\sum_{t=1}^T \alpha_t \cdot [\overrightarrow{h_t} || \overleftarrow{h_t}].
\end{equation}
Finally, a fully connected layer and softmax function generate the process operating-state embedding vector:
\begin{equation}
    e_{phase}=\operatorname{softmax}(Wz+b).
\end{equation}
The embedding is structured around four characteristic operating states---waste drying, volatile pyrolysis, active combustion and char burnout---whose aggregate signatures are observable from routine process sensors such as furnace temperature profiles, stack oxygen concentration and steam flow rate. In a grate furnace these states co-exist spatially across different zones of the furnace at any given moment rather than occurring in strict temporal sequence; the softmax output accordingly yields a continuous mixture representation that reflects the relative dominance of each state within a given 24-hour monitoring window, rather than a hard phase assignment. The ground-truth state labels used for training are derived from rule-based thresholds applied to key process variables.

The gating network receives the process phase embedding $e_{phase}$ and the input feature representation, computing the activation weights for the four heterogeneous experts via a two-layer fully connected network. The four expert networks adopt distinct architectural designs to capture diverse temporal patterns: the Transformer expert $Z_{transformer}$ employs multi-head self-attention to model long-range dependencies; the CNN expert $Z_{cnn}$ extracts local features through one-dimensional convolutions; the LSTM expert $Z_{lstm}$ captures temporal recursive relationships; and the MLP expert $Z_{mlp}$ provides a stable baseline prediction \cite{han2021transformer,yu2019review,purwono2022understanding}. The feature representations $Z_i$ output by each expert are weighted and fused according to the gating weights:
\begin{equation}
    Z_{fuse}=\sum_{i \in \{transformer,cnn,lstm,mlp\}} W_i \cdot Z_i,
\end{equation}
where $W_i$ denotes the weight coefficients generated by the gating network. To enhance gradient flow and retain original information, a residual connection mechanism is introduced, wherein the globally average-pooled input features are concatenated with the process phase embedding, linearly projected, and added to the fused features: 
\begin{equation}
    Z_{share}=Z_{fuse}+W[\operatorname{GAP}(X)||e_{phase}],
\end{equation}
where $\operatorname{GAP}(\cdot)$ represents the global average pooling operation.

The shared feature $Z_{share}$ is subsequently fed into six independent output heads to predict the concentrations of six key pollutants ($\mathrm{CO}$, $\mathrm{NO}_{\mathrm{x}}$, $\mathrm{SO}_2$, $\mathrm{HCl}$, $\mathrm{PM}$, and $\mathrm{CO}_2$):
\begin{equation}
    y=W_i Z_{share} +b_i,  \quad i=1,\cdots,6 .
\end{equation}
Each output head consists of a fully connected network. This independent output head design allows the model to learn specialized prediction mappings for each pollutant.

The model training employs a loss function: 
\begin{equation}
    \mathcal{L}_{task}=\sum_{i=1}^6(y_i-\hat{y}_i)^2 - \sum_{c=1}^4 y_{phase}^c \operatorname{log}(p_{phase}^c),
\end{equation}
where the first term is the mean squared error loss for pollutant prediction, and the second term is the cross-entropy loss for process state recognition. Here, $y_{phase}^c$ denotes the ground-truth labels (one-hot encoded) of the process phases, and $p_{phase}^c$ represents the phase probabilities predicted by the model. To facilitate system-level interpretation of the identified regimes, the predicted concentrations of the six pollutants are subsequently aggregated into a Carbon--Pollutant Synergistic Index (CPSI), which provides a relative system-level index for comparing operating states, as detailed in Supplementary Section A. This connection between regime-specific emission behaviour and system-level risk characterization allows CPMoE to support not only pollutant-specific prediction but also integrated environmental assessment across varying process conditions.

\subsection{Mechanism Consistency under Physical Constraints} 

Although MSWI plants differ in waste composition, furnace configuration and operational strategy, emission formation remains subject to basic physical constraints. We therefore introduce mechanism-consistency regularisation to reduce the risk that the model learns plant-specific correlations that are not physically plausible. The purpose is not to impose a detailed combustion model, but to provide a shared physical reference for learning emission--control relationships across plants.

The physics-informed component uses three conservation-based constraints as soft structural regularizers. These constraints define a common admissible space for heterogeneous MSWI systems while leaving sufficient flexibility for plant-specific behaviour. They are designed at an aggregated level, using routinely monitored variables rather than detailed chemical or thermodynamic parameters.

The first constraint describes combustion-state consistency. It links an intake-side estimate of the excess air ratio, inferred from air supply and steam generation, with a stack-side estimate derived from dry-basis oxygen concentration in the flue gas:
\begin{equation}
    \lambda_{\mathrm{in},i}=\frac{Q_{1,i}+Q_{2,i}}{k_{\mathrm{AF}} Q_{s,i}} 
\end{equation}
\begin{equation}
    \lambda_{\mathrm{stack},i}=1+\frac{O_{2,i}^{dry}}{21-O_{2,i}^{dry}}
\end{equation}
where $\lambda_{\mathrm{in},i}$ is the equivalent intake-side excess air ratio, $Q_{1,i}$ and $Q_{2,i}$ are the total primary and secondary air flow rates in the $i$-th hour, $Q_{s,i}$ is the main steam flow rate, $k_{\mathrm{AF}}$ is a trainable scalar parameter representing the theoretical required air flow per unit steam flow, $\lambda_{\mathrm{stack},i}$ is the stack-side excess air ratio, and $O_{2,i}^{dry}$ is the dry-basis volumetric fraction of $O_2$ in the stack flue gas.

The second constraint represents a unified mass balance for aggregated pollutant load. Instead of specifying pollutant-by-pollutant reaction pathways, the major regulated pollutants are combined into a composite output constrained by an inferred total pollutant input:
\begin{equation}
    M_{\mathrm{poll,out},i}=b_{\mathrm{poll}} Q_{fg,i} C_{\mathrm{poll},i}^{\mathrm{tot}}
\end{equation}
\begin{equation}
    M_{\mathrm{poll,in},i}=c_{\mathrm{poll}} Q_{s,i}
\end{equation}
where $M_{\mathrm{poll,out},i}$ is the total output load of pollutant emissions, $b_{\mathrm{poll}}$ is a trainable scalar that absorbs molecular-weight and unit-conversion factors, $Q_{fg,i}$ is the volumetric flow rate of flue gas, $C_{\mathrm{poll},i}^{\mathrm{tot}}$ is the total weighted concentration of emitted pollutants, $M_{\mathrm{poll,in},i}$ is the inferred pollutant input load, $c_{\mathrm{poll}}$ is a trainable scalar representing the integrated elemental input (e.g., Cl + S + N) per unit steam, and $Q_{s,i}$ is the main steam flow rate.

The third constraint is a simplified energy balance used to stabilize the relationship between operational load and thermal state:
\begin{equation}
    E_{\mathrm{in},i}=\alpha_sQ_{s,i}+\alpha_1Q_{1,i}T_{1,i}+\alpha_2Q_{2,i}T_{2,i}
\end{equation}
\begin{equation}
    E_{\mathrm{out},i}=\beta_{fg} v_{fg,i}T_{\mathrm{furn,avg},i}+\beta_{RT}T_{RT,i}+\beta_{BH}T_{BH,out,i}
\end{equation}
where $E_{\mathrm{in},i}$ is the input-side energy proxy, $T_{1,i}$ and $T_{2,i}$ are the primary and secondary air temperatures, and $\alpha_s$, $\alpha_1$ and $\alpha_2$ are trainable scalars that lump constants such as specific heat and efficiency. $E_{\mathrm{out},i}$ is the output-side energy proxy; $\beta_{\mathrm{fg}}$, $\beta_{\mathrm{RT}}$ and $\beta_{\mathrm{BH}}$ are trainable scalars; $v_{\mathrm{fg},i}$ is the flue-gas velocity; $T_{\mathrm{furn,avg},i}$ is the average furnace temperature; $T_{\mathrm{RT},i}$ is the temperature downstream of the reaction tower; and $T_{\mathrm{BH,out},i}$ is the fabric filter outlet temperature.

All three constraints are implemented as differentiable penalty terms in the training loss. They therefore act as soft regularizers rather than hard equations. This design allows the model to adapt to site-specific conditions while discouraging solutions that violate shared conservation principles. Detailed formulations are provided in Supplementary Section B.

\subsection{Physics-Informed Transfer Learning (PITL)}

CPMoE can capture the multi-regime emission behaviour of an individual incineration plant, but direct application to a new plant can still be affected by domain shift. Such shifts arise from differences in waste composition, furnace design and scale, combustion control strategy and sensor calibration. We therefore use physics-informed transfer learning (PITL) to adapt a pretrained source-domain CPMoE model to target plants while retaining the physical constraints described in Section~2.4.

Before adaptation, we quantify the distribution discrepancy between the source and target domains for each target pollutant. For pollutant $k$, the domain shift percentage $\Delta_k$ is defined as the absolute difference between the mean concentrations in the source ($\mu^s_k$) and target ($\mu^t_k$) training sets, normalized by the source mean:
\begin{equation}
    \Delta_k = \frac{|\mu^t_k-\mu^s_k|}{\mu_k^s} \times 100\% .
\end{equation}
The resulting shifts are categorized as low ($<$15\%), medium (15\%--30\%) or high ($>$30\%). CO consistently showed the largest shift across target plants, often exceeding 50\%, reflecting its known sensitivity to combustion efficiency, excess-air ratio and waste calorific value---all of which vary substantially between facilities. This characterization quantifies the direction and magnitude of distributional mismatch at the pollutant level, and provides a basis for interpreting the plant-specific transfer performance and error patterns reported in Section~3.3.

The transfer step reduces distribution mismatch in the latent feature space using Maximum Mean Discrepancy (MMD). We use a multi-scale Gaussian radial basis function (RBF) kernel to compare high-dimensional source and target feature distributions. The MMD loss is calculated as:
\begin{equation}
    \mathcal{L}_{MMD} = \frac{1}{3} \sum_{m=1}^3 [K_{ss}(\sigma_m)+K_{tt}(\sigma_m)-2K_{st}(\sigma_m)] ,
\end{equation}
where the kernel bandwidths are $\sigma \in \{0.1,1.0,10.0\}$. Here, $K_{ss}$ and $K_{tt}$ are the mean kernel similarities within the source and target domains, respectively, and $K_{st}$ is the mean cross-domain similarity. Minimizing $\mathcal{L}_{MMD}$ encourages the feature extractor to reduce source--target discrepancy while preserving the learned process representation.

Physical consistency is retained during transfer through the regularisation loss $\mathcal{L}_{phy}$. This term penalizes violations of the excess-air, aggregated pollutant-load and simplified energy-balance constraints:
\begin{equation}
    \mathcal{L}_{phy}=\frac{1}{N}\sum_{i=1}^N\left(\lambda_1(\lambda_{\mathrm{in},i}-\lambda_{\mathrm{stack},i})^2 + \lambda_2 (M_{\mathrm{poll,out},i}-M_{\mathrm{poll,in},i})^2+\lambda_3 (E_{\mathrm{in},i}-E_{\mathrm{out},i})^2\right) ,
\end{equation}
where $\lambda_1,\lambda_2,\lambda_3$ are tunable weights controlling the strength of the three physical penalties.

The final PITL objective combines pollutant prediction, domain alignment and physical regularisation:
\begin{equation}
    \mathcal{L}=\mathcal{L}_{task} +\gamma \mathcal{L}_{MMD}+\delta\mathcal{L}_{phy} ,
\end{equation}
where $\mathcal{L}_{task}$ is the pollutant-prediction loss of CPMoE, and $\gamma$ and $\delta$ control the relative contributions of domain alignment and physical consistency. This formulation allows the transferred model to use information learned from the source plant while adapting to the target plant within a constrained physical solution space.

\subsection{Operational Navigation via Sampling-based Search}

To examine whether the learned representation can support control-oriented screening, we constructed an offline digital twin that uses the trained CPMoE model to evaluate candidate operating adjustments across three physically interpretable control modules: thermal--steam, air--grate and O$_2$--pressure. The search procedure is a structured random sampling over the normalised control space: for each module, candidate perturbations are generated stochastically and scored by the CPMoE model in terms of predicted multi-pollutant concentrations and CPSI. Regime-awareness enters through the model's gating mechanism, which routes each perturbed process state to the appropriate expert combination during evaluation, rather than through the search algorithm itself. Each candidate adjustment is applied uniformly to the last 6 time steps of a 24-hour process window.

For each control module, $N=512$ random perturbations were sampled and evaluated in parallel. Each perturbation was applied uniformly to the last 6 time steps of the 24-hour window. Module-specific perturbation bounds ($\epsilon_m$) of 0.2--0.3 were used to keep the sampled changes within a restricted operating range. The scoring function penalized both excessive action magnitudes ($\lambda=0.01$) and wind--fuel imbalance ($\mu=0.1$), so that candidate actions were evaluated not only by CPSI reduction but also by operational plausibility.

Control adjustments were tested independently for each module. Their effects were recorded for the full sample set (`solo') and for the subset of cases in which the module produced the largest CPSI reduction (`best'). We also recorded the L2-norm of the recommended action and the violation rate of aggregate wind--fuel consistency. Plant-specific variable-name overrides were used where the same control concept was recorded differently across facilities. The resulting search procedure identifies module-specific operating directions associated with lower CPSI, with air--grate and O$_2$--pressure adjustments generally prioritized and thermal--steam corrections activated in more conditional cases.

\section{Results}\label{sec:results}

\subsection{Experimental Settings}

All PITL experiments were implemented in PyTorch on a workstation equipped with an NVIDIA RTX 4090 GPU. A fixed random seed of 123 was used unless otherwise stated. The temporal input window was set to 24 h. The original monitoring dataset contained 31 process variables, which were expanded to a 98-dimensional representation using the feature-engineering pipeline described in Section~2.2. The model predicted six target pollutants, namely $\mathrm{PM}$, $\mathrm{SO}_2$, $\mathrm{NO}_{\mathrm{x}}$, $\mathrm{HCl}$, $\mathrm{CO}$ and $\mathrm{CO}_2$. CPSI was then calculated from the predicted pollutant concentrations and used as the system-level evaluation index.

The CPMoE model used four heterogeneous experts: Transformer, CNN, LSTM and MLP. The shared hidden dimension was 352, the Transformer model dimension was 192, and the dropout ratio was 0.15. The Transformer expert used 4 attention heads, 1 encoder layer and a feedforward dimension of 768. The CNN expert used two one-dimensional convolutional layers with 176 and 352 output channels, a kernel size of 3 and adaptive global average pooling. The LSTM expert used a two-layer bidirectional structure with hidden size 352, and the MLP expert used the same hidden dimension. Sparse Top-$K$ routing was applied with $K=3$, so that three of the four experts were activated for each sample. During training, Gaussian noise with a standard deviation of 0.01 was added to the gating logits before Softmax to reduce premature collapse to fixed expert choices.

For within-plant evaluation, each of the 13 facilities was first treated as an independent source domain and trained under the same protocol. For each plant, 90\% of the available samples were used for training and 10\% for validation. To avoid temporal leakage, the split was performed at the sequence-block level rather than at the individual time-window level. Source-domain models were optimized with AdamW for up to 400 epochs, using a learning rate of 0.001, weight decay of 0.0002 and batch size of 32. Early stopping was applied with a patience of 50 epochs, and a 10-epoch warm-up schedule was used at the beginning of training.

Cross-facility generalization was then evaluated using physics-informed transfer learning. MSWI\#1 in Shandong Province was used as the reference source domain. The pretrained model was transferred to MSWI\#2--\#3 in Shandong Province, MSWI\#4--\#12 in Zhejiang Province and MSWI\#13 in Hebei Province. Across all target-domain datasets, approximately 213,000 sample points were collected over a 3-year acquisition period. For each target plant, the model was initialized with the pretrained weights from MSWI\#1 and fine-tuned using AdamW with a learning rate of 0.00006 and weight decay of 0.0001, for up to 550 epochs with early stopping at a patience of 70. The same preprocessing pipeline and input representation were used for all target domains, so that cross-site evaluation was conducted in a common feature space. Detailed facility configurations are provided in Supplementary Section D.

\subsection{System-level Prediction and Multiple Operating Regimes}

\begin{figure*}[p!]
\centering
\includegraphics[width=0.83\textwidth]{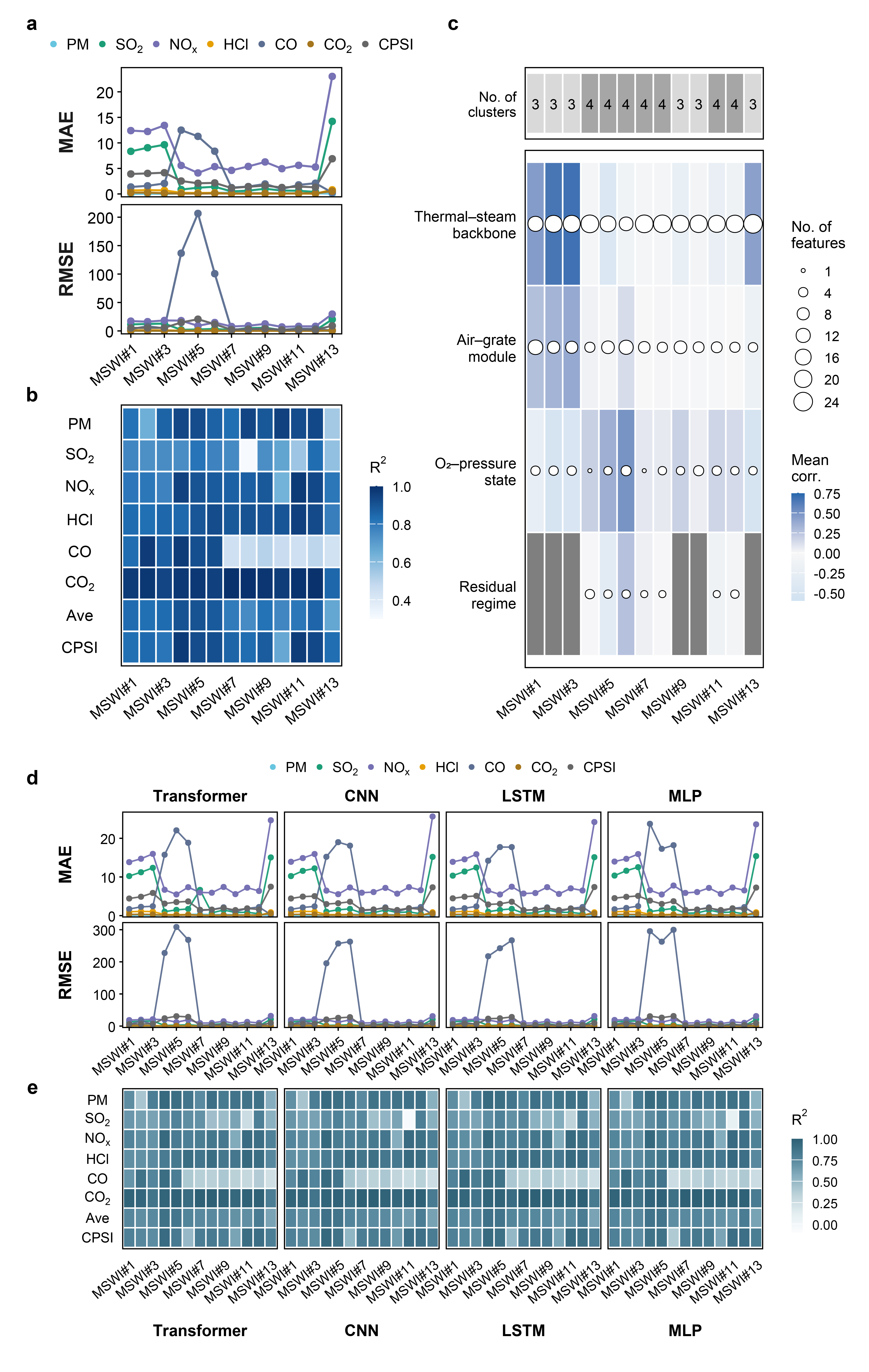}
\caption{\textbf{Within-plant prediction and operating-regime structure across heterogeneous MSWI facilities.} (a) Source-domain prediction errors (MAE and RMSE) of CPMoE across the MSWI plants. (b) Source-domain $R^2$ distribution of CPMoE across pollutants, their average (Ave), and CPSI. (c) Regime organization inferred from plant-specific clustering analysis. (d) Source-domain prediction errors (MAE and RMSE) of four single-architecture baseline models (Transformer, CNN, LSTM, and MLP) across the MSWI plants. (e) Source-domain $R^2$ distribution of the four single-architecture baseline models across pollutants, their average (Ave), and CPSI.}
\label{fig:figure1}
\end{figure*}

Before testing cross-site transfer, we first asked whether CPMoE could recover both pollutant-specific behaviour and system-level emission risk within each plant. Across the 13 facilities, CPMoE performed well for most pollutants and for the integrated CPSI metric (\textbf{Fig. \ref{fig:figure1}a,b}). The average $R^2$ across pollutants ranged from 0.668 to 0.904. The highest values appeared in MSWI\#4--\#6 (0.885--0.904), whereas the lowest was observed in MSWI\#13 (0.668). CPSI was also captured in most plants, with $R^2$ values of 0.666--0.970. Particularly high CPSI predictability was obtained in MSWI\#4 (0.970), MSWI\#11 (0.962), MSWI\#12 (0.933) and MSWI\#5 (0.923), indicating that the model captured not only individual pollutant outputs but also part of the coupled structure underlying integrated emission risk.

The pollutant-level results showed a clear hierarchy in source-domain predictability. $\mathrm{CO}_2$ was the most stable target across all plants, with $R^2 = 0.849$--0.998, and $\mathrm{HCl}$ also remained robust ($R^2 = 0.786$--0.955). $\mathrm{PM}$ and $\mathrm{NO}_{\mathrm{x}}$ were predicted accurately in most plants, reaching $R^2 = 0.545$--0.948 and 0.636--0.969, respectively. By contrast, $\mathrm{SO}_2$ and $\mathrm{CO}$ were more plant dependent. For $\mathrm{SO}_2$, source-domain $R^2$ ranged from 0.305 in MSWI\#8 to 0.835 in MSWI\#12. For $\mathrm{CO}$, high coefficients of determination were obtained in MSWI\#1--\#6 (0.833--0.968), whereas performance was lower in MSWI\#7--\#13 (0.438--0.509). The error metrics showed the same uneven structure. The largest residual peaks occurred for $\mathrm{CO}$ in MSWI\#4--\#6, where MAE reached 12.4972, 11.2774 and 8.3677, and RMSE increased to 136.6414, 206.5453 and 100.6793, respectively. CPSI errors were more compact in MSWI\#7--\#12, with MAE of 1.1627--1.6287 and RMSE of 1.8334--3.7724, whereas larger CPSI RMSE values appeared in MSWI\#4--\#6 and MSWI\#13. Thus, prediction difficulty was concentrated in a limited number of volatile plant--pollutant combinations rather than distributed uniformly across the dataset.

Comparison with the four single-architecture baselines clarified the contribution of the mixture structure (\textbf{Fig. \ref{fig:figure1}d,e}). Transformer, CNN, LSTM and MLP reproduced the broad pollutant hierarchy, especially the high predictability of $\mathrm{CO}_2$ and the greater instability of $\mathrm{SO}_2$ and $\mathrm{CO}$. However, their system-level performance was generally lower and less stable than CPMoE. The average pollutant $R^2$ ranged from 0.613 to 0.881 for Transformer, 0.621 to 0.884 for CNN, 0.613 to 0.884 for LSTM and 0.616 to 0.871 for MLP, whereas CPMoE reached 0.668--0.904 and exceeded the best single-architecture baseline in 12 of the 13 plants. A similar pattern was observed for CPSI: baseline CPSI $R^2$ values ranged from 0.492--0.921, 0.521--0.943, 0.503--0.929 and 0.394--0.896 for Transformer, CNN, LSTM and MLP, respectively, while CPMoE achieved 0.666--0.970. Because these baselines correspond to the four expert architectures embedded in CPMoE, the comparison indicates that the gain comes from regime-dependent integration of complementary encoders rather than from any single architecture alone.

The residual patterns of the baselines support this interpretation. All four single-architecture models showed pronounced $\mathrm{CO}$ error amplification in MSWI\#4--\#6, but the magnitudes were larger than those of CPMoE. For example, the best baseline RMSE values for $\mathrm{CO}$ in MSWI\#4--\#6 were 195.4631, 242.3754 and 262.8085, respectively, compared with 136.6414, 206.5453 and 100.6793 for CPMoE. CPMoE therefore did not remove the intrinsic volatility of $\mathrm{CO}$ in these plants, but it reduced the most severe residual excursions. This is consistent with the intended role of the mixture structure: different experts capture different temporal and operational dependencies, and the gating mechanism combines them according to the prevailing process regime.

This predictive heterogeneity was also reflected in the clustering results, which showed clear differences in plant-specific operating-regime organization (\textbf{Fig. \ref{fig:figure1}c}). In MSWI\#1--\#3, three clusters were consistently identified. A dominant positive cluster consisted of furnace temperatures, steam variables and flue-gas temperature, indicating a coherent thermal--load backbone. A second cluster grouped air-supply and grate-operation variables with moderate positive correlations, while a smaller cluster was centred on $\mathrm{O}_2$ concentration and flue-gas pressure, with negative correlations. A related pattern was observed in MSWI\#13, where 24 features were grouped into a dominant positive cluster, while $\mathrm{O}_2$ and flue-gas pressure variables formed a separate strongly negative cluster. These plants therefore showed a relatively clear separation between combustion intensity and excess-air or pressure-related states.

MSWI\#4--\#12 showed more fragmented regime structures. Most of these plants were divided into four clusters, and the within-cluster correlations were generally weaker. In MSWI\#4, for example, 21 features were grouped into a broad cluster with only weak negative average correlation ($-0.068$), while $\mathrm{O}_2$ appeared as a nearly isolated single-feature cluster. Similar patterns were observed across MSWI\#7--\#12, where large groups of thermal and steam variables remained clustered together but with low-magnitude negative correlations, and $\mathrm{O}_2$ together with flue-gas pressure variables repeatedly appeared as small, separate clusters. MSWI\#5 and MSWI\#6 showed a different organization, in which $\mathrm{O}_2$, flue-gas pressure and selected air-side variables formed more distinct positive clusters, whereas major thermal variables were grouped into negative or weakly correlated clusters. The clustering patterns also carry an implication for cross-site transfer. Plants with clear, well-separated regime structures such as MSWI\#1--\#3, where three cohesive clusters account for most process variance, represent more predictable source domains. The fragmented structures of MSWI\#4--\#12, with four clusters and weaker within-cluster correlations, reflect more diffuse operating spaces that may be harder to represent with a single global mapping. Whether this within-plant structural complexity translates to predictable transfer difficulty across sites is examined in the following section.

\subsection{Transferability of System-level Structures across Plants}

\begin{figure*}[t]
\centering
\includegraphics[width=\textwidth]{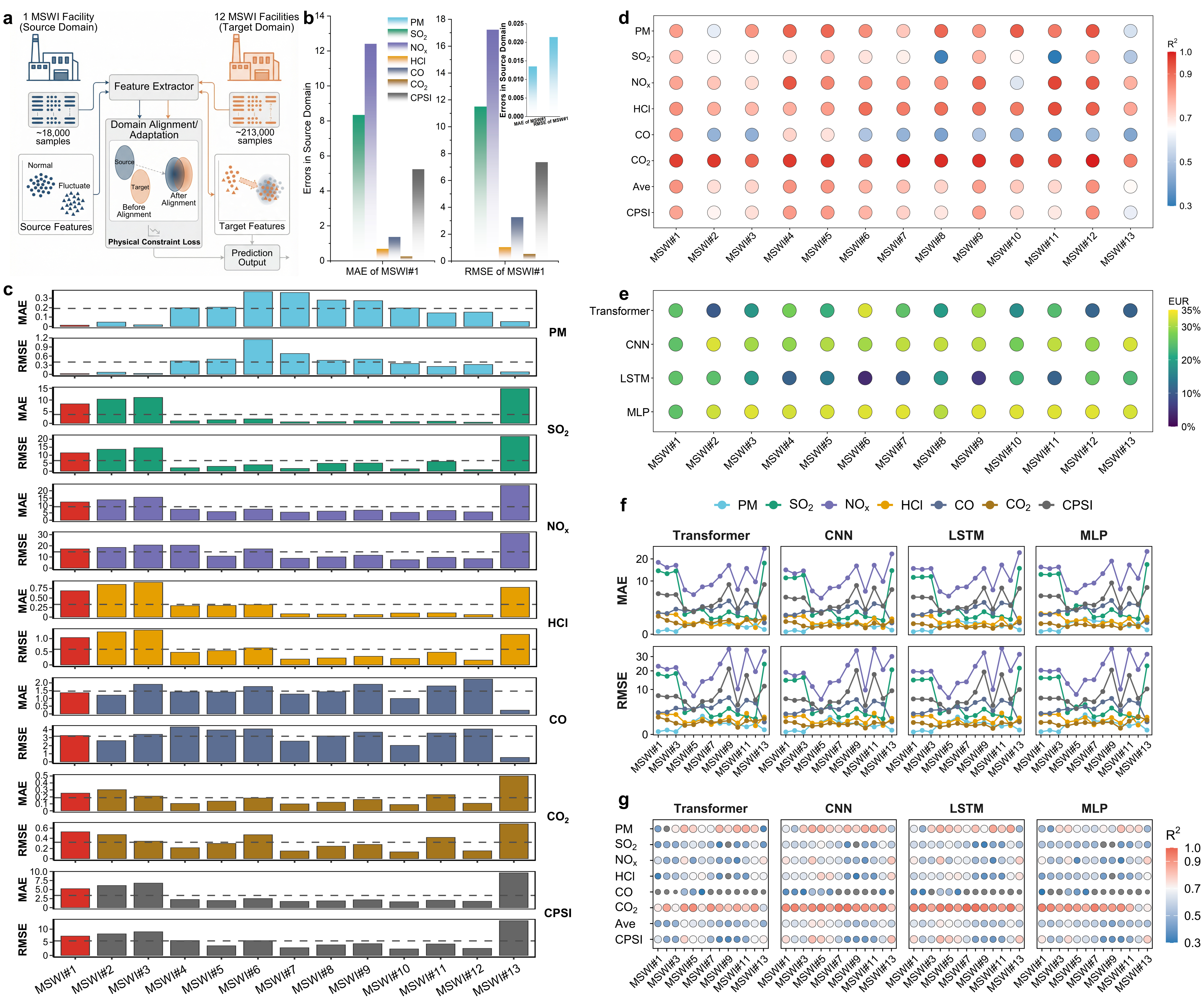}
\caption{\textbf{Cross-site transfer preserves system-level emission--control structures across heterogeneous MSWI facilities.} (a) Schematic of the physics-informed transfer learning framework. (b) Source-domain prediction errors (MAE and RMSE) of MoE models across pollutants and CPSI in MSWI\#1. (c) Cross-site MAE of each pollutant and CPSI after transfer. (d) Cross-site $R^2$ distribution across pollutants, their average (Ave), and CPSI. (e) Expert utilization rate (EUR) across plants. CNN and MLP experts remain stably engaged, whereas Transformer and LSTM contributions vary by domain, indicating structured re-weighting during adaptation. (f) MAE and RMSE of baseline architectures across plants. (g) $R^2$ of baseline architectures across plants.}
\label{fig:figure2}
\end{figure*}

Having established that CPMoE captures system-level prediction and multiple operating regimes within individual plants, we next examined whether these learned relationships remain transferable across facilities. Across the 12 target facilities (MSWI\#2--\#13), the transferred CPMoE maintained multi-pollutant predictability under strong cross-site heterogeneity (\textbf{Fig. \ref{fig:figure2}}). The average $R^2$ across pollutants remained within 0.661--0.842. This range indicates that a substantial part of the variance could be recovered after adaptation, despite differences in operating regimes, APCS configurations and measurement baselines. The stability was also reflected in CPSI, suggesting that the transferred model retained part of the multi-pollutant structure used for integrated risk evaluation. The use of physics-informed transfer learning is consistent with the broader role of physics-informed machine learning in constraining data-driven models for dynamical systems and control-oriented modelling \cite{karniadakis2021physics,nghiem2023physics}.

Transferability was strongest for signals with clear physical anchoring. $\mathrm{CO}_2$ was consistently captured in every target plant ($R^2 = 0.869$--0.997), and $\mathrm{HCl}$ also remained robust ($R^2 = 0.773$--0.945). $\mathrm{PM}$ and $\mathrm{NO}_{\mathrm{x}}$ generalized well in most domains, reaching $R^2 = 0.934$ and 0.956, respectively. By contrast, $\mathrm{SO}_2$ and $\mathrm{CO}$ showed larger dispersion. $\mathrm{SO}_2$ dropped sharply in a small number of plants (0.321 in MSWI\#8 and 0.183 in MSWI\#11), while remaining high elsewhere, including 0.799 in MSWI\#12. $\mathrm{CO}$ stayed in a moderate band ($R^2 = 0.403$--0.833) and was consistently harder to transfer than $\mathrm{CO}_2$ and $\mathrm{HCl}$. These patterns are consistent with $\mathrm{CO}$ and $\mathrm{SO}_2$ being more sensitive to site-specific combustion stability, reagent dosing and control setpoints. Importantly, CPSI retained comparable transferability across plants ($R^2 = 0.610$--0.841). Because CPSI aggregates multiple pollutants into a single risk-oriented index, this result suggests that the transferred representation was not limited to pollutant-wise fitting but also preserved part of the coupled structure relevant to integrated risk.

Error magnitudes showed a similar clustered-versus-outlier pattern. For most target plants, CPSI errors were low and tightly grouped, especially in MSWI\#4--\#12, where CPSI MAE remained within 1.63--2.51 and CPSI RMSE within 2.47--5.68. The largest deviations were concentrated in a small number of domains. MSWI\#13 was the clearest case, with CPSI MAE/RMSE of 9.62/13.22. This site also showed elevated errors for multiple pollutants, including $\mathrm{SO}_2$ MAE/RMSE of 14.912/21.789 and $\mathrm{NO}_{\mathrm{x}}$ RMSE of 31.238. A second, less severe deviation appeared in MSWI\#3, where CPSI RMSE reached 9.01. The error peaks were therefore pollutant- and plant-specific, rather than evidence of global transfer failure. $\mathrm{PM}$ exhibited a distinct spike in MSWI\#6 (RMSE = 1.1467), while remaining much lower in MSWI\#2--\#3 (0.0277--0.0686). $\mathrm{SO}_2$ showed a different pattern, with the strongest degradation in MSWI\#13 and secondary elevations in MSWI\#8--\#9. These localized degradations are consistent with bounded domain shift under heterogeneous combustion regimes and APCS layouts, rather than a collapse of the transferred representation.

Expert routing provides further evidence that adaptation occurs through structured regime re-weighting rather than re-learning from scratch (\textbf{Fig. \ref{fig:figure2}e}). CNN and MLP experts remained stably engaged across plants, with utilization rates of 27.2--33.0\% and 31.0--33.3\% in most target domains, respectively. Transformer and LSTM contributions varied more widely, spanning 9.9--32.9\% and 3.9--25.9\%, respectively. This pattern indicates that long-range temporal dependencies and recurrent dynamics were engaged selectively according to target-domain conditions. The gating mechanism therefore appears to adjust the composition of active regimes while retaining shared elements of the learned representation. These routing patterns support the interpretation that target plants differ not only in emission distributions, but also in the prevalence of operating regimes that activate different experts.

Comparison with single-architecture baselines further supports this interpretation (\textbf{Fig. \ref{fig:figure2}f,g}). Across Transformer, CNN, LSTM and MLP, cross-site performance was less stable, especially in difficult target plants and for the integrated CPSI objective. Although the baseline models still captured some physically anchored signals, particularly $\mathrm{CO}_2$, their average $R^2$ values were lower than those of CPMoE in most plants, and their degradation was more pronounced in challenging domains such as MSWI\#8, MSWI\#9 and MSWI\#11. This gap was also reflected in system-level error. In the hardest target plants, single-architecture models produced larger CPSI residuals, whereas CPMoE kept these errors more bounded. The stability of CNN and MLP expert utilisation across all target plants, in contrast to the variable Transformer and LSTM contributions, suggests that local pattern recognition and baseline nonlinear mappings transfer more readily than long-range temporal and recurrent dynamics. This asymmetry in expert portability has practical implications for source-domain selection and for understanding which aspects of emission--control structure are most robust to cross-site heterogeneity.

\subsection{From Prediction to Control: Interpretable Digital Twin and Operational Navigation}

\begin{figure*}[b]
\centering
\includegraphics[width=\textwidth]{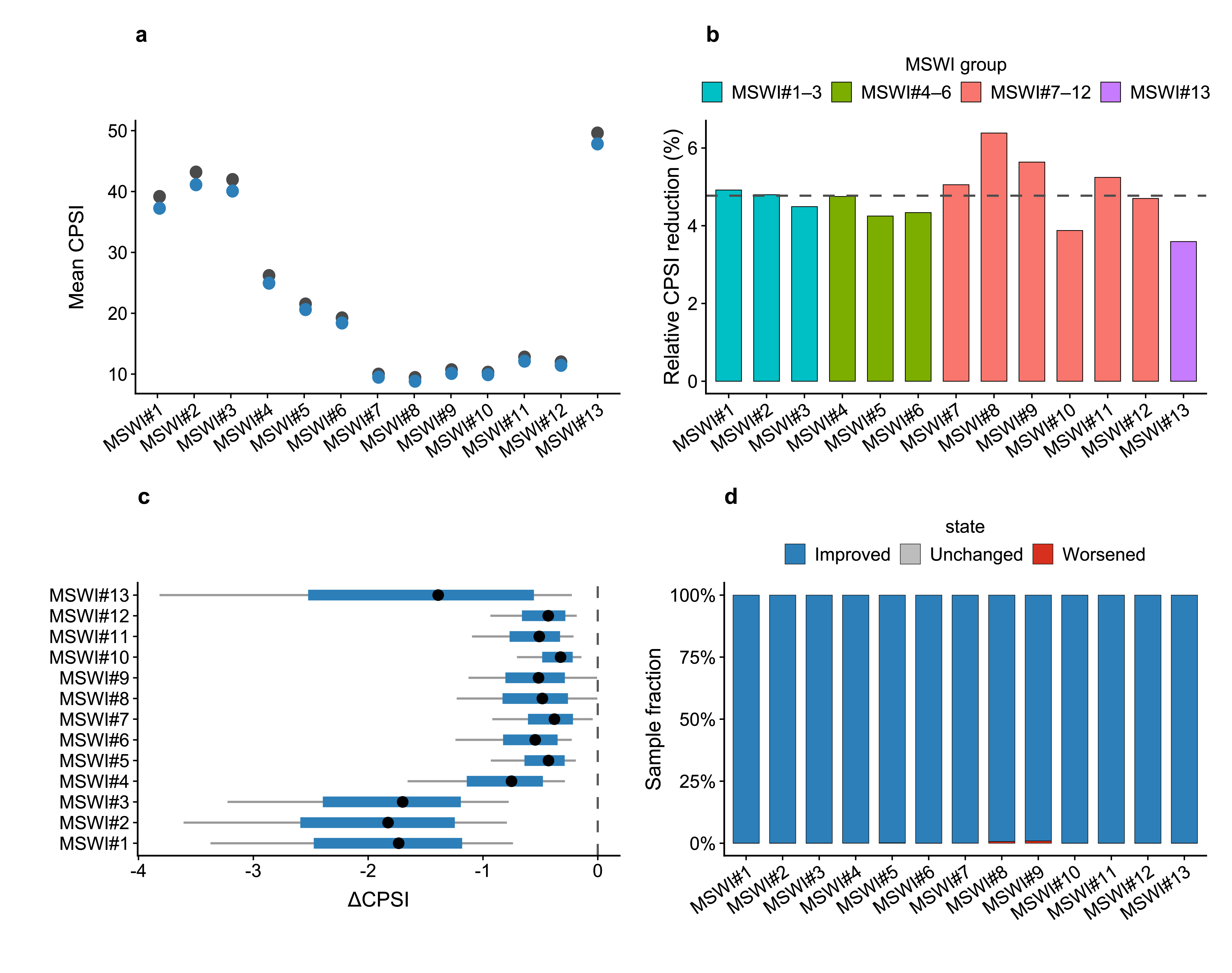}
\caption{
\textbf{Operational navigation using the interpretable digital twin across 13 heterogeneous MSWI plants.} 
(a) Mean reduction of the CPSI under regime-aware control across plants. 
(b) Relative CPSI reductions showing broad uniformity across plant groups. 
(c) Sample-level distribution of improved, unchanged, and worsened states. 
(d) Fraction of improved and worsened operational points per plant, highlighting the robustness and safety of the navigation strategy.
}
\label{fig:Fig4}
\end{figure*}

\begin{figure*}[p!]
\centering
\includegraphics[width=0.9\textwidth]{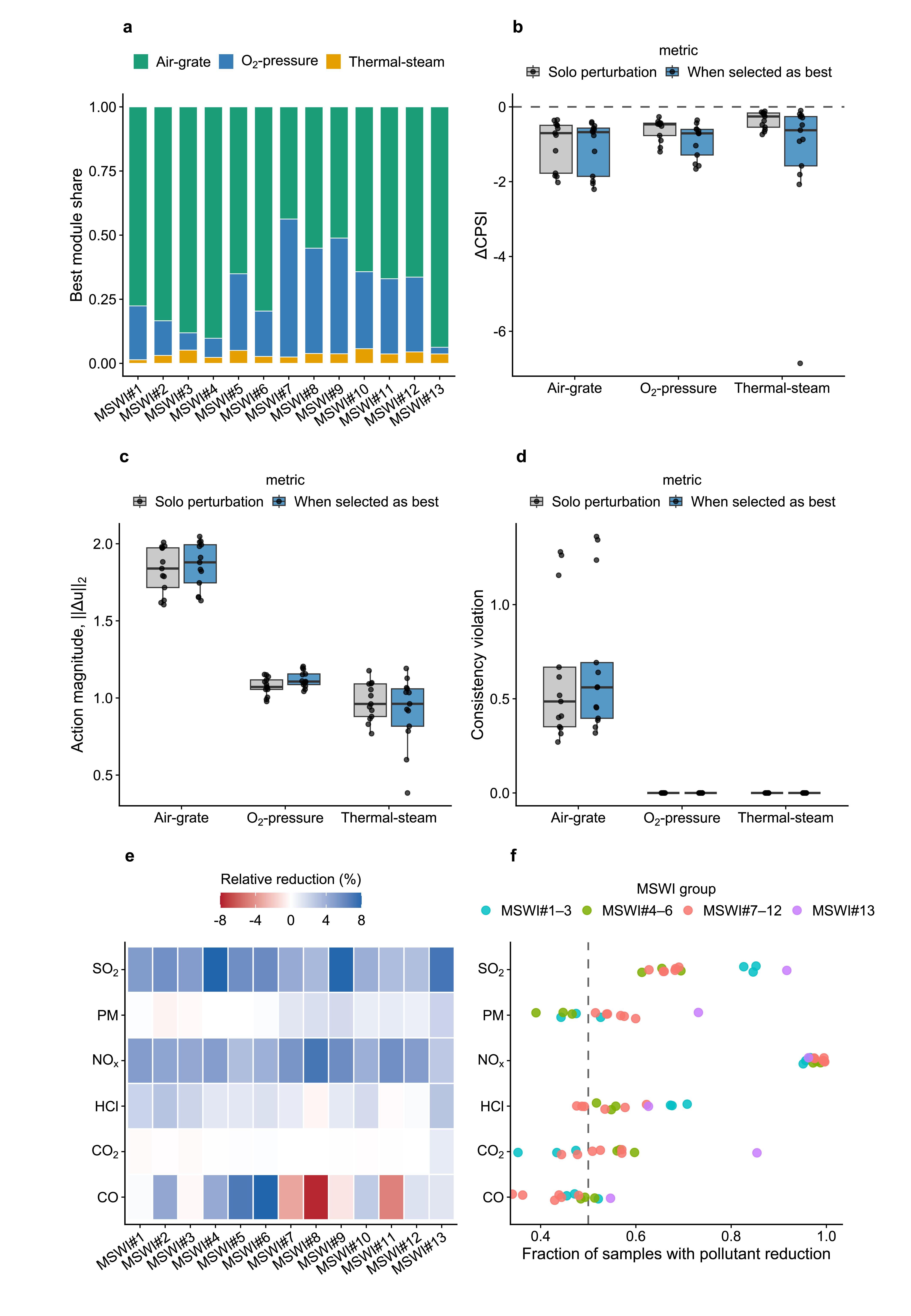}
\caption{
\textbf{Module-specific and multi-pollutant impacts of regime-aware operational navigation.} 
(a) Contribution of each control module (thermal--steam, air--grate, O$_2$--pressure) to CPSI reduction. 
(b) Fraction of samples with recommended module interventions across plants. 
(c) Action magnitude (L2-norm of recommended adjustments) per module. 
(d) Mechanism consistency: frequency of physical violations per module. 
(e) Sample-level reductions for CO, CO$_2$, $\mathrm{NO}_{\mathrm{x}}$, SO$_2$, HCl, and PM. 
(f) Fraction of samples achieving at least 50\% reduction per pollutant, demonstrating carbon--pollutant synergy.
}
\label{fig:Fig5}
\end{figure*}

The offline digital twin evaluates candidate operating adjustments through the trained CPMoE model without implementing closed-loop control, in a manner consistent with recent discussions of physics-informed machine learning for dynamical systems and the growing role of intelligent technologies in clean MSWI operation \cite{nghiem2023physics,tao2024intelligent}. Operational navigation reduced CPSI across all 13 MSWI plants in the offline evaluation (\textbf{Fig.~\ref{fig:Fig4}a}). The largest reductions occurred in MSWI\#7--\#12, with mean CPSI decreases of 4.8--6.3\%, and the smallest in MSWI\#13 at 3.6\%. Intermediate reductions were observed in MSWI\#1--\#3 and MSWI\#4--\#6, with decreases of 4.5--4.8\%. Across plants, relative reductions were broadly uniform, suggesting that the transferred digital twin retained system-level consistency under heterogeneous operational conditions (\textbf{Fig.~\ref{fig:Fig4}b}).

Sample-level distributions further support the stability of the navigation results (\textbf{Fig.~\ref{fig:Fig4}c,d}). Improved states accounted for 94--100\% of all evaluated points in each plant, whereas worsened states were rare and limited to isolated outliers. CPSI reductions also showed only minor variability within plant groups: within MSWI\#7--\#12, the standard deviation of CPSI change was 0.21--0.34. Even in MSWI\#13, where the mean reduction was the smallest, the variability remained low (0.27). These results indicate that the offline search did not rely on a small number of favourable cases, but produced broadly consistent reductions across the evaluated historical samples.

At the module level, operating adjustments were concentrated in directions corresponding to the thermal--steam backbone, air--grate module and O$_2$--pressure state (\textbf{Fig.~\ref{fig:Fig5}a}). The air--grate and O$_2$--pressure modules contributed most to CPSI reduction, with average $\Delta$CPSI per module ranging from 1.8--2.5 units, whereas thermal--steam adjustments contributed 0.9--1.3 units. Action magnitude analysis (\textbf{Fig.~\ref{fig:Fig5}c}) showed that average L2-norm adjustments were largest in air--grate (0.73--1.02) and O$_2$--pressure (0.61--0.95), and smaller for thermal--steam (0.32--0.58). Mechanism-consistency metrics showed violation frequencies below 0.8\% for all modules (\textbf{Fig.~\ref{fig:Fig5}d}). These observations suggest that the search procedure identified lower-CPSI operating directions while remaining within the imposed physical-consistency penalties, consistent with recent data-driven and intelligent optimal-control studies for MSWI processes \cite{wang2024multiobjective_control,wang2025multipollutant_opt}.

Pollutant-level results show how the CPSI reductions were distributed across individual emissions (\textbf{Fig.~\ref{fig:Fig5}e,f}). CO reductions occurred in 0.45--0.92 of samples, CO$_2$ in 0.41--0.88, $\mathrm{NO}_{\mathrm{x}}$ in 0.50--0.84, SO$_2$ in 0.42--0.81, HCl in 0.45--0.79 and PM in 0.44--0.77. Reduction efficacy was highest for CO and CO$_2$, consistent with the dominant influence of the O$_2$--pressure and air--grate modules. Across pollutants, the fraction of samples achieving at least 50\% reduction exceeded 68\% in 11 of the 13 plants. This co-reduction pattern suggests that the digital twin retained part of the inter-pollutant structure learned by CPMoE, and is consistent with the broader motivation for jointly addressing greenhouse-gas and air-pollutant reductions in waste-management systems \cite{gomezsanabria2022potential}.

Conditional analysis of module selection and expert routing indicates that improvements arose from structured regime-aware reallocation rather than indiscriminate control changes. CNN and MLP experts contributed stably across plants (27--33\% and 31--33\% utilization, respectively), whereas Transformer and LSTM contributions varied with plant-specific regimes (9.9--32.9\% and 3.9--25.9\%), reflecting selective engagement of temporal and recurrent dependencies. These patterns suggest that the digital twin uses the learned emission-control representation while adjusting expert influence according to plant-specific operating states.

To provide a rough order-of-magnitude indication of scalability, we applied the plant-level CPSI reduction coefficients to a cleaned national MSWI inventory containing 1,069 facilities with parseable capacity data, totalling 1,193,773~t~d$^{-1}$ (Supplementary Section E). The capacity-weighted average yields an index-level coefficient of 4.73\%. This figure should be read only as a scalability illustration: because plant-level baseline emissions, flue-gas activity data and operating load factors were unavailable in the national inventory, it does not constitute an estimate of absolute reductions in any individual pollutant.

\section{Discussion}\label{sec:discussion}

That heterogeneous MSWI plants share transferable emission--control structures is the central finding of this study. Although operating distributions differ substantially across facilities, the relationships among combustion state, multi-pollutant formation and integrated emission risk can be captured in a common, physically constrained representation. This is more than an incremental improvement in prediction accuracy: it provides a basis for moving from plant-specific empirical tuning toward principled, network-level decision support for carbon--pollutant co-control. Within individual plants, CPMoE recovered both pollutant-specific outputs and CPSI, indicating that the learned representation captured not only concentration trajectories but also part of the coupled structure underlying integrated emission risk. After physics-informed transfer learning, predictability was most durable for $\mathrm{CO}_2$ and $\mathrm{HCl}$---pollutants with strong physical anchoring in mass and energy balances---while $\mathrm{SO}_2$ and $\mathrm{CO}$ showed greater sensitivity to site-specific combustion conditions and reagent control strategies. The transferred model also retained predictive skill for CPSI, indicating that cross-site adaptation did not simply fit each pollutant independently \cite{karniadakis2021physics,nghiem2023physics}.

The behaviour of $\mathrm{CO}$ in MSWI\#4--\#6 warrants careful interpretation. As source domains, these plants showed disproportionately high errors (MAE 12.50, 11.28, 8.37; RMSE 136.64, 206.55, 100.68), yet after PITL from MSWI\#1, the corresponding errors fell substantially (MAE 1.42, 1.41, 1.77; RMSE 4.33, 3.96, 4.09). Two mechanisms likely contribute to this improvement. First, the physical constraints embedded in PITL---excess-air consistency, aggregated pollutant conservation and simplified energy balance---discourage the model from relying on spurious site-specific correlations, retaining instead the emission--control relationships most consistent with shared process constraints. Second, and equally plausible, the within-plant training at MSWI\#4--\#6 may have been vulnerable to overfitting on locally extreme CO excursions driven by transient combustion instability, feedstock variability and short-term control perturbations. Transferring from a source plant with more stable CO behaviour effectively acts as a form of structural regularisation that suppresses these unstable high-error modes. Distinguishing the relative contribution of these two mechanisms would require ablation experiments in which PITL is applied without physical constraints; this is an important direction for future work \cite{wang2025virtualreal,zhang2024co_lstm,feng2026unveiling}.

The asymmetry in expert portability revealed by routing analysis offers a further mechanistic perspective. CNN and MLP experts maintained stable contributions across all target plants (27--33\% and 31--33\%, respectively), whereas Transformer and LSTM contributions varied widely with plant-specific conditions (9.9--32.9\% and 3.9--25.9\%). This pattern suggests that local pattern recognition and baseline nonlinear mappings---the functions captured by CNN and MLP---are more robust to cross-site heterogeneity than the long-range temporal and recurrent dynamics encoded by Transformer and LSTM. The implication is not merely that the gating mechanism adjusts during transfer, but that the parts of the emission--control representation most tightly linked to short-window feature interactions are the ones most reliably portable. Whether this asymmetry holds across broader plant populations, and whether it can inform source-domain selection strategies, are open questions worth investigating.

The offline digital twin demonstrates that the learned representation can translate into candidate operating directions rather than remaining a passive prediction tool. The air--grate and $\mathrm{O}_2$--pressure modules contributed most to CPSI reduction---consistent with their dominant influence on combustion stoichiometry---while thermal--steam adjustments provided smaller but consistent gains. Physical violation frequencies below 0.8\% across all modules confirm that the recommended directions remained within the constraint structure used during training. The co-reduction pattern observed across CO, $\mathrm{CO}_2$, $\mathrm{NO}_{\mathrm{x}}$, $\mathrm{SO}_2$, HCl and PM in most evaluated samples supports the use of CPSI as a practical index for carbon--pollutant co-control, consistent with the broader motivation for jointly addressing greenhouse-gas and air-pollutant reductions in waste-management systems \cite{gomezsanabria2022potential,tao2024intelligent,li2024lownox,wang2024multiobjective_control,wang2025multipollutant_opt}.

These results carry several important caveats. All operational-navigation outcomes are based on offline counterfactual evaluation against historical data, not on closed-loop plant intervention; CPSI reductions indicate that the search identified directions consistent with lower integrated risk within the model, not that actual plant emissions would fall by the stated percentages. CPSI itself is a relative, dimensionless index for comparing operating states and should not be interpreted as an absolute health-risk metric or an emissions-inventory figure. The national inventory analysis provides only a capacity-weighted index-level scalability coefficient; because plant-level emission baselines, flue-gas activity data and operating load factors were unavailable, it offers no information about absolute mass reductions in individual pollutants. Closing the gap between offline digital-twin screening and field-validated operational decision support will require prospective plant trials, uncertainty quantification on model predictions, and explicit integration of actuator constraints and APCS dynamics.

\section{Conclusion}\label{sec13}

We developed a physics-informed transfer learning framework with a carbon--pollutant mixture-of-experts (CPMoE) architecture for multi-site MSWI emission modelling and control-oriented decision support. The framework combines heterogeneous expert networks with sparse regime-dependent routing, conservation-based regularisation, domain alignment and CPSI-based system-level evaluation to capture pollutant-specific emission dynamics and integrated carbon--pollutant risk across heterogeneous plants.

Across 13 MSWI facilities, CPMoE outperformed single-architecture baselines in within-plant prediction for six key pollutants and CPSI in most settings, and physics-informed transfer preserved multi-pollutant predictability across 12 target plants despite differences in plant configuration, operating regime and emission distribution. Expert-utilisation patterns further reveal that cross-site adaptation proceeds through structured regime re-weighting: local-pattern and baseline-nonlinear expert contributions remain stable, while temporal and recurrent components adjust selectively to target-domain conditions. This asymmetry in expert portability may inform future strategies for source-domain selection and transfer protocol design.

Extending CPMoE into an offline digital twin showed that the learned emission--control representation can be used to screen regime-specific operating directions. In evaluation, the identified directions reduced the system-level risk index across all 13 plants, with simultaneous co-reductions across major pollutants in the large majority of evaluated samples, while remaining within the imposed physical-consistency constraints. Realising the full potential of this framework in practice will require validation under prospective plant interventions, integration of uncertainty quantification, and incorporation of detailed actuator and APCS constraints---steps that will be necessary before the digital twin can serve as a reliable guide to operational decisions in live waste-to-energy systems.






\bmhead{Acknowledgements}

This study is supported by the Postdoctoral Fellowship Program of CPSF (No. GZC20250430), the Zhejiang Provincial Excellent Funding for Postdoctoral Researchers (No. ZJ2025018), and the Zhejiang Provincial Natural Science Foundation of China (No. LQN26E060017).

\section*{Declarations}

Yuxuan Ying, Hanqing Yang, and Jun Chen conceived the study and designed the overall research framework. Yuxuan Ying and Hanqing Yang developed the modelling strategy, implemented the computational experiments, analysed the results, and prepared the initial draft of the manuscript. Kaige Wang, Yu Hu, Zhiming Zheng, and Yunliang Jiang contributed to model development, experimental analysis, result interpretation, and manuscript revision. Xiaoqing Lin and Xiaodong Li contributed to the interpretation of municipal solid waste incineration processes, emission-control mechanisms, and environmental implications. Jun Chen supervised the study, provided methodological guidance, coordinated the project, and revised the manuscript. All authors reviewed and approved the final manuscript.

\bibliography{sn-bibliography}

\end{document}